\documentclass{article}

\usepackage[final]{corl_2023} 

\usepackage{graphics} 
\usepackage{epsfig} 
\usepackage{mathptmx} 
\usepackage{times} 
\usepackage{amsmath} 
\usepackage{amssymb}  
\usepackage{siunitx} 
\usepackage{textcomp}
\usepackage{gensymb} 
\usepackage{booktabs}
\usepackage{multirow}
\usepackage{algpseudocode}
\usepackage{algorithm}
\usepackage{diagbox}
\usepackage{lipsum}
\usepackage{wrapfig}

\let\labelindent\relax
\usepackage{enumitem}
\usepackage{xspace}
\setlist{nolistsep}

\usepackage{makecell}
\usepackage{multicol}
\usepackage{rotating}
\usepackage[percent]{overpic}
\usepackage{contour}
\usepackage{courier}

\usepackage{subcaption} 
\usepackage[font=small]{caption}
\usepackage[percent]{overpic}

\definecolor{MyDarkRed}{rgb}{0.8,0.02,0.02}
\definecolor{MyDarkGreen}{rgb}{0.02,0.6,0.02}

\definecolor{MyPurple}{rgb}{0.6,0.1,.9}

\newcommand{\rebuttal}[1]{#1}
\newcommand{\algo}{SpawnNet}

\usepackage[toc,page,header]{appendix}
\usepackage{minitoc}

\title{SpawnNet: Learning Generalizable Visuomotor Skills from Pre-trained Networks}
\author{
  Xingyu Lin$^*$ \, 
  John So$^*$ \, 
  Sashwat Mahalingam \,
  Fangchen Liu \,
  Pieter Abbeel \\[2mm]
  UC Berkeley \\[2mm] 
  $^*$Equal Contribution
}

\begin{document}
\doparttoc 
\faketableofcontents 
\maketitle


\begin{abstract}
The existing internet-scale image and video datasets cover a wide range of everyday objects and tasks, bringing the potential of learning policies that generalize in diverse scenarios. Prior works have explored visual pre-training with different self-supervised objectives. Still, the generalization capabilities of the learned policies and the advantages over well-tuned baselines remain unclear from prior studies. In this work, we present a focused study of the generalization capabilities of the pre-trained visual representations at the categorical level. We identify the key bottleneck in using a frozen pre-trained visual backbone for policy learning and then propose SpawnNet, a novel two-stream architecture that learns to fuse pre-trained multi-layer representations into a separate network to learn a robust policy. Through extensive simulated and real experiments, we show significantly better categorical generalization compared to prior approaches in imitation learning settings. Open-sourced code and videos can be found on our website: \url{https://xingyu-lin.github.io/spawnnet/}.

\end{abstract}



\begin{figure}[ht]
  \centering
  \includegraphics[width=\linewidth]{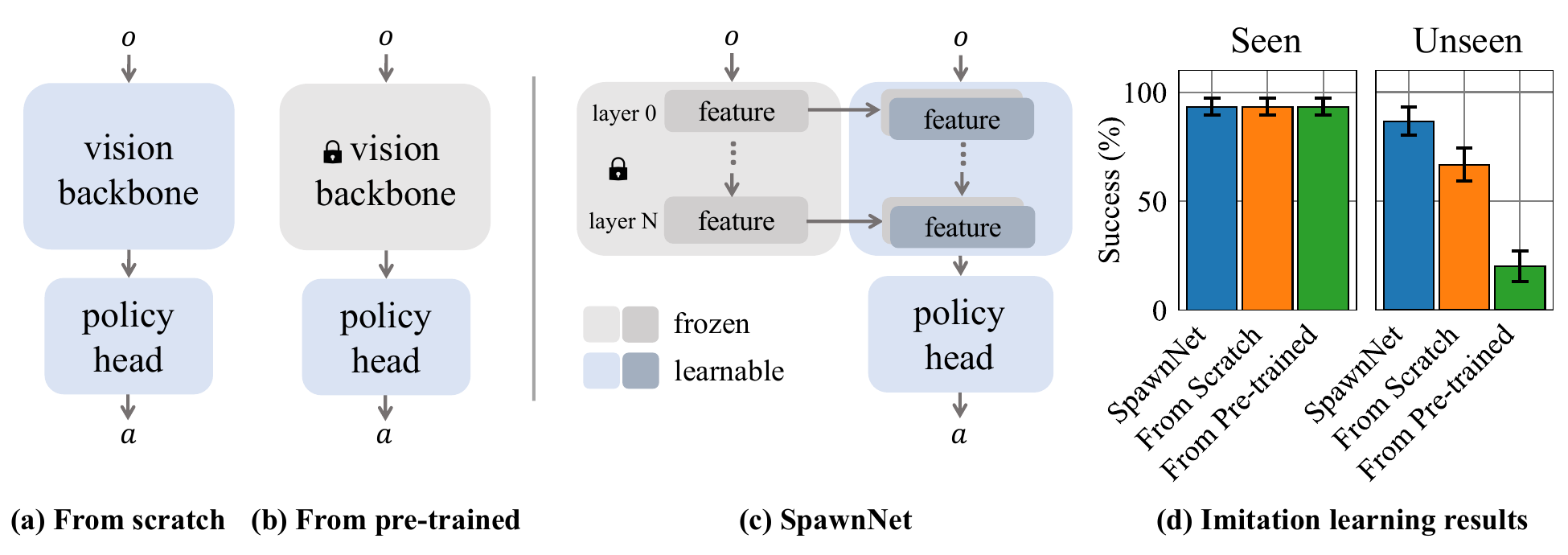}
  \caption{Prior approaches for learning policies (a) from scratch, (b) from a pre-trained visual representation with a frozen backbone, and (c) the proposed two-stream architecture. The right figure (d) shows their performances on a real-world imitation learning task, evaluated on both seen and unseen instances in a category.}
  \label{fig:pull}
\end{figure}

\section{Introduction}
To take steps towards deploying robots in our daily lives, learning skills that can handle diverse situations is crucial for robots to subsist with us in the real world. For example, consider a scenario where we want our robot to help us wear hats. We can teach this skill by demonstrating the robot with the hats we currently have. However, we don't want to recollect demonstrations once we buy a new hat or change the position of our coat rack. Thus, we hope to enable robots with a smart hat-wearing policy that can handle object variations in color, pose, and shape.


To this end, we need to equip manipulation skills with semantic knowledge of the world, which already lies in current internet-scale video and image datasets~\cite{damen2018scaling,rauman2022ego4d,schuhmann2022laion}. Given the success of visual representation learning~\cite{he2020momentum, grill2020bootstrap, he2022masked, chen2020simple, chicco2021siamese, oquab2023dinov2} on a wide range of computer vision tasks, a natural solution to our goal is to take the pre-trained representations out of the box and train robust visuomotor skills on top of it. Prior works have explored this direction~\cite{radosavovic2022real,nair2022r3m,ma2022vip,karamcheti2023language}, with promising results on improved sample efficiency and asymptotic performance. However, most evaluations in prior works are done on simple tasks with relatively small variations and thus are not sufficiently difficult. Indeed, recent work~\cite{hansen2022pretraining} has shown that a carefully done baseline that learns from scratch can be very competitive with these prior works that use pre-trained networks.



\begin{figure}[ht!]
    \centering
    \includegraphics[width=\linewidth]{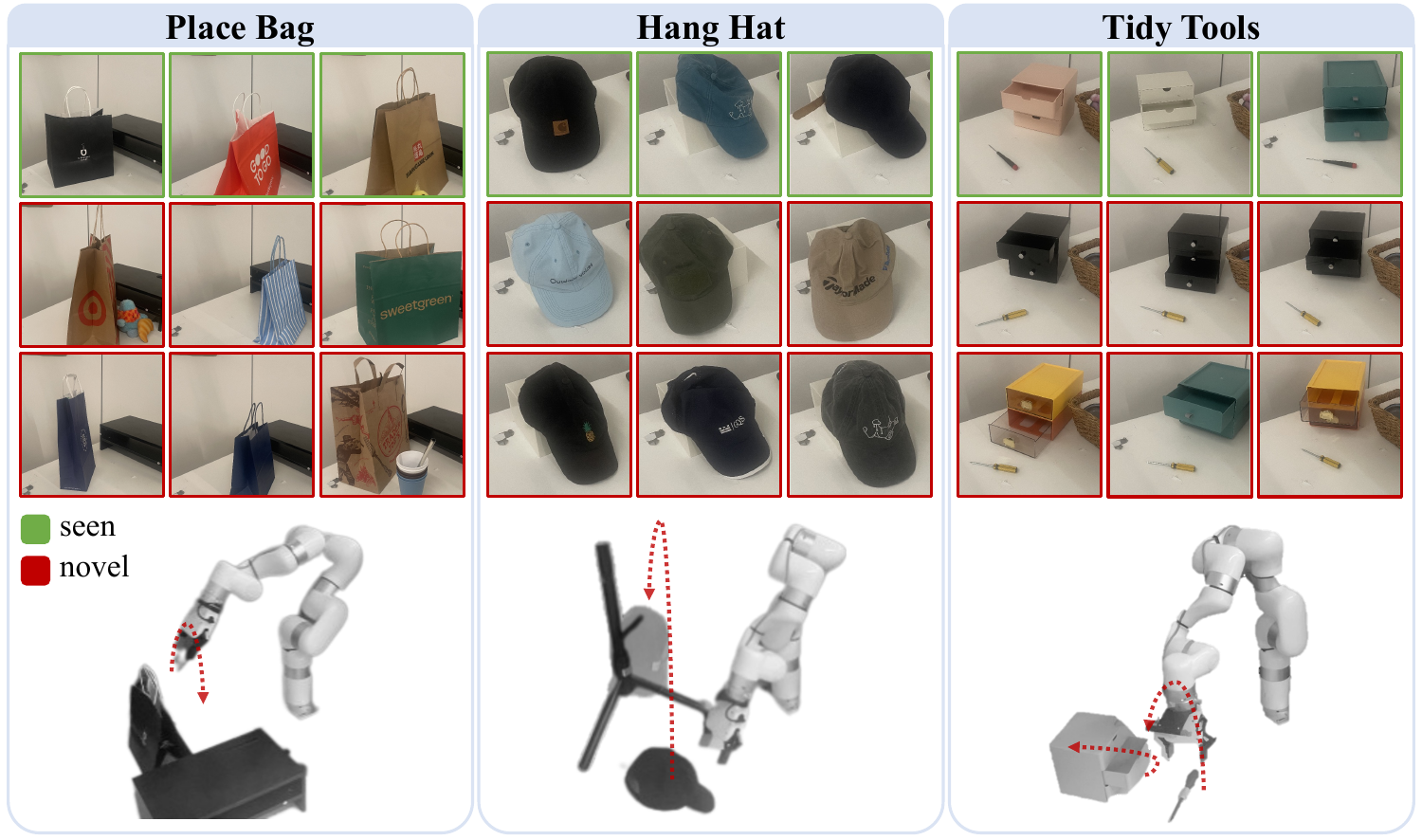}
    \caption{We consider three challenging categorical manipulation tasks in the real world. For each task, we train on three instances (green boxes) and test on held-out instances (red boxes), with additional variations in poses, articulation, visual distraction, and deformation.}
    \label{fig:real_tasks}
\end{figure}

We hypothesize that the benefits of pre-training will be more pronounced in more complex generalization tests, where a trained policy is evaluated for the ability to adapt its behavior to unseen objects. Hence, we create three challenging categorical manipulation tasks (depicted in Figure~\ref{fig:real_tasks}). Our tasks contain a set of diverse objects with no overlapping instances between training and evaluation. Through these challenging tasks, we identify the bottleneck in the existing methods: frozen pre-trained networks give fixed representations to the policy and as such, may hinder policy learning without adapting the visual backbone. The need for adaptation may come from the differences between the pre-training and policy learning objectives, or from the distribution shift from the pre-training dataset to robotic demonstrations, including domains, tasks, or camera viewpoints~\cite{zhao2022distshift}.

Inspired by the ProgressiveNet~\cite{rusu2016progressive} in the continual learning literature, we propose a simple two-stream architecture, \algo, to learn a generalizable visuomotor policy from a pre-trained neural network. Instead of directly using frozen representations, we adopt a separate network that learns to fuse multi-layer pre-trained representations (Figure~\ref{fig:pull}) and generate actions. The learnable stream incorporates domain-specific features from the raw observations to help handle distribution shifts. Meanwhile, it can take advantage of the pre-trained features at different layers for faster learning and better generalization. While other transfer learning methods have shown successes in CV and NLP~\cite{houlsby2019parameter,hu2021lora,chen2022vision}, including simple fine-tuning, they can underperform in our case due to the large discrepancy between visual pre-training and downstream control tasks. Through extensive experiments, we demonstrate that {\algo } significantly outperforms prior approaches and a learning-from-scratch method when tested on held-out novel objects. Below we highlight the contribution of this paper:
\begin{enumerate}[topsep=1.0pt,itemsep=1.0pt,leftmargin=5.5mm]
    \item We propose~\algo, a simple and flexible framework that can adapt any pre-trained model to a generalizable visuomotor policy on downstream manipulation tasks.
    \item We perform systematic evaluations of different methods utilizing pre-trained representation both in simulation and in the real world, showing significant improvement of our method in cross-instance generalization.
\end{enumerate}

\section{Related Works}
\textbf{Visual Pre-training for End-to-end Policy Learning}. Recently there has been a surge of works that aim to pre-train a visual representation on image or video datasets, using ResNet~\cite{shah2021rrl, parisi2022unsurprising, nair2022r3m} or Vision Transformer (ViT) \cite{radosavovic2022real, khandelwal2022simple} as the backbone. After training, these works freeze the visual representation for downstream policy learning. It has been shown that pre-trained representation can achieve better sample efficiency or asymptotic performance. However, evaluations in prior works are done in environments with limited visual variations, such as changes in object poses. Some works study changes in backgrounds~\cite{chen2023genaug,yu2023scaling}, which do not require the policy to adapt its action sequences. As such, a recent study shows that strong learning-from-scratch baselines with data augmentations can achieve comparable performance to the best pre-trained representations~\cite{hansen2022pretraining}. In contrast, our work sheds light on the generalization capabilities of pre-trained representations on a rich set of assets and tasks in both simulation and real-world experiments through a novel neural architecture that better utilizes the pre-trained representation. We note concurrent work that uses pre-trained networks in large-scale manipulation tasks~\cite{brohan2022rt}, while our study is more focused and studies the challenging categorical manipulation tasks. 

\textbf{Transfer Learning}. Transferring knowledge from a pre-trained network has led to success in many downstream tasks in CV and NLP, using a frozen representation or fine-tuning as the de facto approach during adaptation. As the pre-trained networks become larger, parameter-efficient fine-tuning approaches (PEFT) have also been proposed by inserting adaptation modules in the pre-trained network~\cite{houlsby2019parameter,pfeiffer2020adapterfusion,hu2021lora,chen2022vision}. However, few works have explored adaptation architectures for vision-based control in robotics. Sharma et al.~\cite{sharma2023lossless} apply the PEFT idea for adapting a pre-trained representation. However, PEFT methods may not have enough capacity to overcome the gap between the visual pre-training and downstream control tasks, as we will show in experiments. Our architecture is similar to Progressive Neural Network~(ProgNet)~\cite{rusu2016progressive} in continual learning, but extends over ProgNet by having asymmetric network architectures for the pre-trained stream and control stream.

\section{Method}

Pre-trained visual representations are trained on large image and video datasets and have the potential to help learn generalizable visuomotor skills. We will first give a background on the vision transformer, the pre-trained vision backbone architecture we use in Section~\ref{sec:vit_prelim}. Then in Section~\ref{sec:arch}, we explain the issue with the current way of using the pre-trained features and present our architecture. Finally, we explain our policy learning frameworks for evaluation in Section~\ref{sec:policy_learning}.

\subsection{Vision Transformer Preliminary} \label{sec:vit_prelim}
While our method can be combined with any pre-trained visual architecture, we will mainly discuss its instantiation with pre-trained vision transformers~\cite{dosovitskiy2020image} as they are more scalable and commonly used for visual pre-training. The input RGB images are first split into patches. Each patch is encoded into a latent vector representation named a token. Additionally, a learned CLS token is concatenated to the other image tokens, and the sequence is passed through multiple transformer layers. Each transformer layer consists of alternating layers of multi-headed self-attention (MHSA) and MLPs, with residual layers between them. Within the MHSA block, an input $X$ is first split, then projected onto learned key, query, and value bases before the attention mechanism: $MHSA(X)=\textit{softmax}(QK^T / \sqrt{d})V$, where $Q, K, V$ are linear projections of $X$.

Prior works in using the pre-trained representation for policy learning usually take the CLS token at the last layer as the image representation. On the other hand, people have found the attention features in MHSA in intermediate layers of the vision transformer to encode semantic information at the object-level and part-level~\cite{amir2021deep,hadjivelichkov2023one}. As such, we follow \cite{amir2021deep} to extract the key $K$ in MHSA from a pre-trained vision transformer DINO~\cite{caron2021emerging}. Since vision transformers usually use a large model and require large computation to fine-tune, throughout this paper we freeze the weights of the pre-trained network.

\begin{wrapfigure}{R}{0.4\textwidth}
  \vspace{-5mm}
  \includegraphics[width=0.4\textwidth]{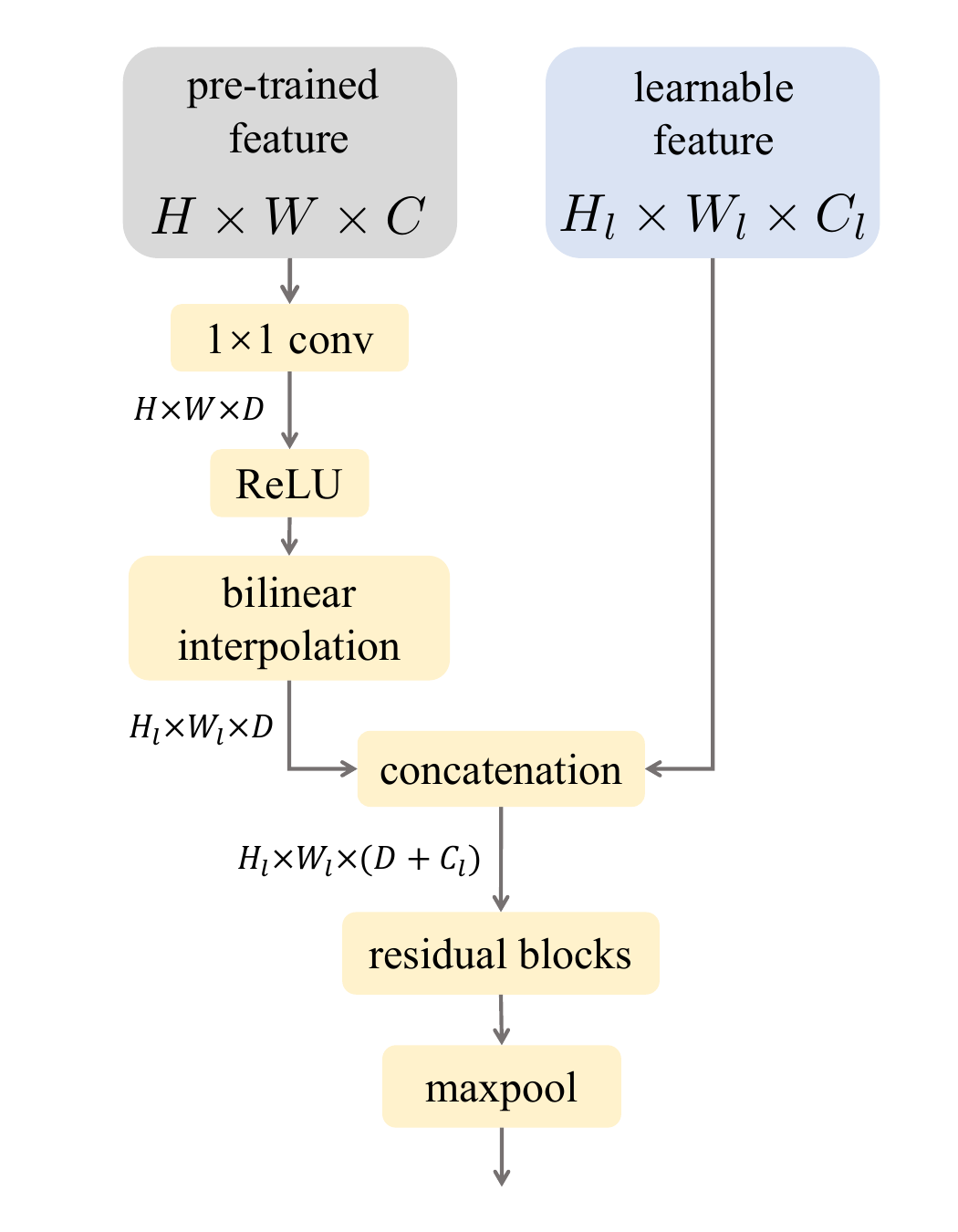}
  \caption{Adapter layers to fuse the pre-trained features with the learnable features.}
  \vspace{-3mm}
  \label{fig:arch}
\end{wrapfigure}

\subsection{\algo~Architecture} \label{sec:arch}
Given these powerful pre-trained vision transformers, how do we utilize them for policy learning? Prior approaches extract the CLS token at the last layer as the image representation, as illustrated in Figure~\ref{fig:pull}. The representation is typically frozen as fine-tuning vision transformers requires large computation. However, this architecture lacks flexibility in adapting the visual features. This potentially hinders downstream policy learning for two reasons. First, the self-supervised objectives in pre-training may not align with the policy learning objective. Second, datasets used for pre-training can have a different distribution from downstream tasks in domains, tasks, or camera viewpoints~\cite{zhao2022distshift}.

To learn more flexible task-specific representations while taking advantage of the pre-trained features, \algo~trains a separate stream of convolutional neural networks~(CNN) from scratch, taking the raw observation as input~(Figure~\ref{fig:pull}). At the same time, we design adapter modules to fuse the new stream with the pre-trained features at different layers, taking advantage of the robust features as needed. We find that a shallow CNN for the new stream works well under the low-data regime during policy learning. Additionally, this architecture allows us to incorporate different modalities like depth in the separate stream, which is not in the pre-trained vision networks.

We first extract spatially dense descriptors from different layers of the pre-trained networks. At the $l^{th}$ layer of the vision transformer, we extract the key feature in MHSA from each token, forming a feature grid of $\phi^{l}(I) \in \mathbb{R}^{H \times W \times C}$, where $H \times W$ is the number of image patches and $C$ is the feature dimension. Extracting features from different layers allows us to extract both low-level and high-level visual features from the pre-trained network~\cite{amir2021deep}. Note that vision transformers keep the same number of image patches at each layer; extracting features from each image patch allows us to maximally preserve spatial information.

\textbf{Adapter Layers}. We fuse the features from the two streams of networks at different layers as shown in Figure~\ref{fig:arch} using \textit{adapter layers}. The adapter first maps the pre-trained feature into a latent embedding of size $D$ through a convolutional kernel of size 1 and stride 1. It then resizes the feature from $H \times W$ to $H_l \times W_l$ with bi-linear interpolation and finally concatenates them in the feature dimension. It is then passed through two residual blocks to be processed in the next layers. \rebuttal{We note that the design of the adapter layers allows us to utilize flexible architectures for downstream control tasks. This is a key difference to ProgressiveNet~\cite{rusu2016progressive} and enables us to effectively learn policies from limited demonstration data. We will demonstrate this in the experiments.}
 
\subsection{Visuomotor Policy Learning} \label{sec:policy_learning}
We study the effect of visual representation in learning challenging visuomotor control tasks. For all our tasks, we consider large intra-category variations and hold out a portion of instances during training to study generalization.

\textbf{Learning with Expert Guidance in Simulation}. Prior works study the visual representations under the Reinforcement Learning~(RL) or Behaviour Cloning~(BC) framework. This couples the visual representation with challenges in exploration or covariate shift. For our simulation experiments, we first train RL policies on all training instances using PPO~\cite{schulman2017proximal}. We treat the RL policies as experts and learn image-based policies using DAgger~\cite{ross2011reduction}: Within each iteration, we first roll out the current policy in the environments and then query the experts on agent's visited states to get the corresponding action labels for training. We then train the current policy with gradient descent updates to minimize the MSE loss between the agent's actions and the expert's actions.

\textbf{Behavior Cloning with Teleoperation Demonstration in the Real World}. While the simulated tasks allow us to iterate quickly, the rendered images are not photo-realistic, which creates an artificial gap to the real data used for pre-training. As such, we further train and evaluate visuomotor policies in the real world. For each task, we collect demonstrations from humans using a teleoperation setup. Since it is difficult to query humans in robot-visited states in the real world, we simply train different policies using behavior cloning.

\section{Experiments}
Below, we provide experiments for (1) comparisons with prior works using pre-trained representations in simulation (Sec.~\ref{sec:sim} and the real world (Sec.~\ref{sec:real_world} (2) visualization of the emergent attention on parts~(Sec.~\ref{sec:policy_learning}) (3) ablation study~(Sec.~\ref{sec:ablation}) (4) SpawnNet with different pre-trained networks~(Sec.~\ref{sec:other-networks} and (5) comparisons with other transfer learning methods~(Sec.~\ref{sec:transfer}). We note that experiments in prior works~(\cite{radosavovic2022real,nair2022r3m,hansen2022pretraining}) have only shown limited variations in poses or instances within a category. We aim to stress-test the generalization capabilities of the policies through significant in-category variations where a policy would break if not well adapted.

\textbf{Baselines}. We compare with the following baselines:
\begin{itemize}[leftmargin=8mm]
    \item \textbf{LfS+aug (Learning from Scratch)~\cite{hansen2022pretraining}} is a shallow ConvNet encoder followed by an MLP policy with data augmentation, which has been shown to be a strong baseline. Following \cite{hansen2022pretraining}, we use random shift as our data augmentation.
    \item \textbf{MVP~\cite{radosavovic2022real}} trains a masked auto-encoder from ego-centric video data and takes the frozen CLS token from the vision transformer as the representation. 
    \item \textbf{R3M~\cite{nair2022r3m}} trains a language-aligned visual representation from videos using time contrastive learning and language-video alignment. 
\end{itemize}
All methods with only a frozen pre-trained backbone (i.e. \textbf{MVP}, \textbf{R3M}) take RGB as input. Meanwhile, incorporating depth is straightforward for \textbf{{\algo}} and \textbf{LfS} by adding a depth channel to the trainable encoder. We denote these depth-conditioned variants as \textbf{\algo+d} and \textbf{LfS+aug+d}.

\subsection{Simulation Experiments} \label{sec:sim}
\begin{figure}[ht]
  \centering
  \includegraphics[width=\textwidth]{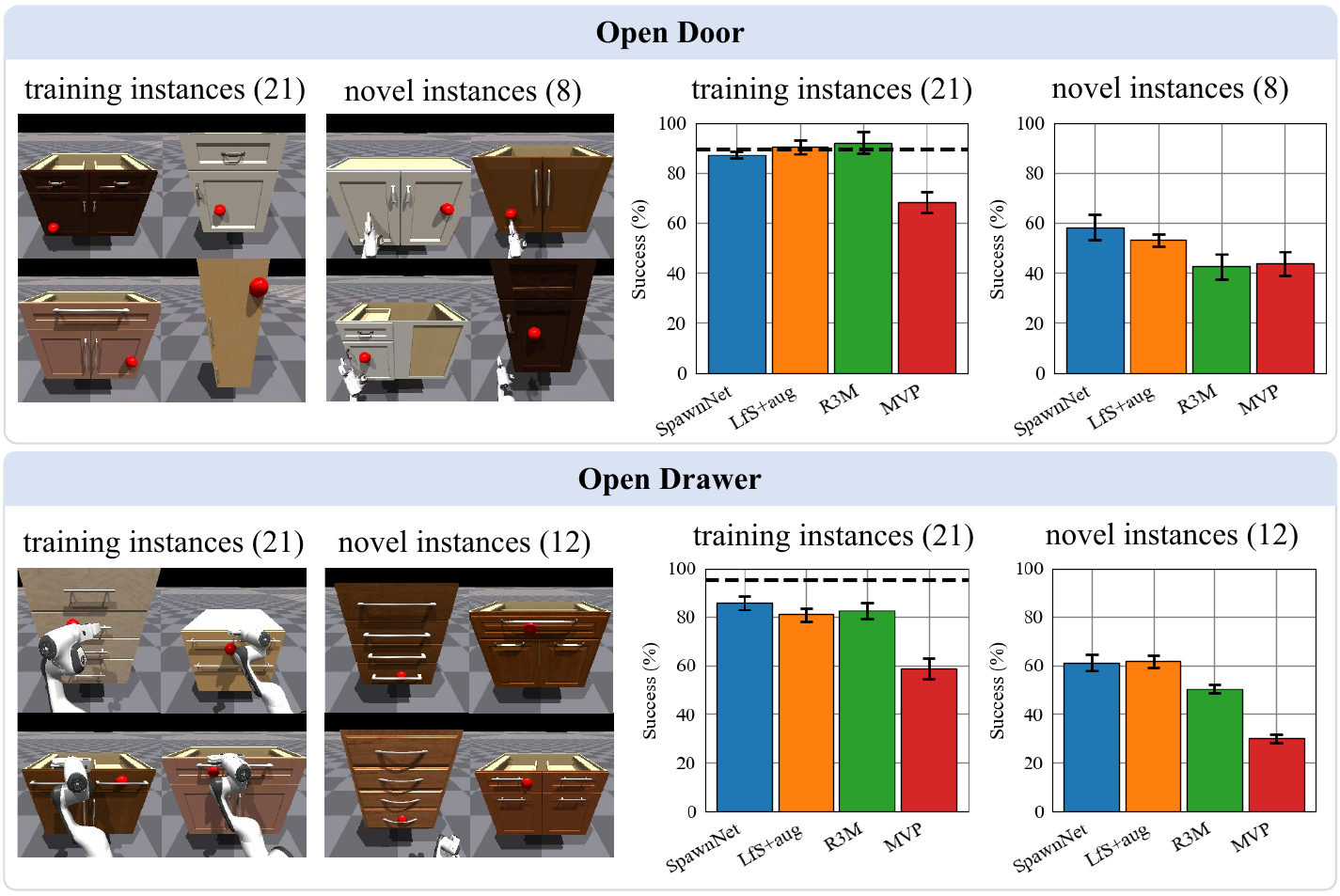}
  \caption{Simulation results on Open Door and Open Drawers. The left figures show the training and novel instances we use. The observations are rendered from the agent's middle camera. We add red spheres in the scene to specify the task of which door/drawer to open. The right shows success rates of different methods in both seen and unseen instances after a fixed number of agent rollouts. The dashed black line shows the RL expert's performance. The error bars show the standard error computed from three random seeds. Numbers in the brackets denote the number of instances. }
  \label{fig:exp_sim}
\end{figure}

\textbf{Tasks}. We conduct experiments on two tasks in simulation, opening different cabinet doors and drawers with a Franka arm, as shown in Figure~\ref{fig:exp_sim}. For each task, we hold out a subset of the objects during training and use them for evaluating the generalization of the learned policies on novel objects. Our tasks are taken from RLAfford~\cite{geng2022end}, built on top of IsaacGym~\cite{makoviychuk2021isaac}. The agents receive image observations from three camera views (left, middle, right) and the proprioception as input, and output joint positions for the arm. 

\textbf{Training and Evaluation}. We first take the pre-trained RL policies from~\cite{geng2022end} using PPO~\cite{schulman2017proximal}, which takes point clouds as input. We then train different policies with DAgger to imitate the RL experts. This helps us decouple the visual representation from other difficulties such as the covariant shift in behavior cloning. We alternate model training and agent rollouts and stop after a fixed number of trajectories. For all methods, we select 21 instances from the training set where the experts perform well and use them for training DAgger policies. We evaluate policies on 8 held-out instances for Open Door and 12 held-out instances for Open Drawer. We roll out the agent for 5 trajectories per asset to evaluate each model. For each method, we train three models with different seeds. We report the average performance and the standard error across training and novel instances. 

The results are shown in Figure~\ref{fig:exp_sim}. We find that different approaches perform similarly in training instances, while SpawnNet generalizes better than other pre-trained representations. LfS+aug proves to be a strong baseline in simulation. We next evaluate in a real-world setting, where visual pre-training should comparatively benefit from realistic image observations and limited training instances.

\subsection{Real World Experiments} \label{sec:real_world}
\textbf{Experimental Setup}. We evaluate methods using a real-world robotics setup with a UFACTORY xArm 7. Humans provide demonstrations using a 3Dconnexion Spacemouse; actions are parameterized and collected as 6-DOF, delta end-effector control $(\Delta x, \Delta y, \Delta z, \Delta \alpha, \Delta \beta, \Delta \gamma)$ at 5Hz, plus gripper open/close. Both translation and rotation in the action space are defined in the wrist camera view to enable better generalization. Each demonstration collects observations as RGB and depth images from two RealSense cameras: one from third-person view, and one attached to the wrist of the robot arm. We stack the most recent four frames as the agents' observations. We do not give the agents access to proprioception information as we find that the agent tends to overfit to the proprioception and ignore the visual input, which hinders generalization.

For real-world comparison, we evaluate against \textbf{R3M} and \textbf{LfS+aug+d} as they perform well in simulation tasks. 

\textbf{Tasks}. We evaluate methods with behavior cloning for three manipulation tasks. Each real-world task consists of around 90 demonstrations across 3 training instances with pose variations and a set of held-out objects. The tasks and variations are illustrated in Figure~\ref{fig:real_tasks} and summarized below:
\begin{enumerate}[topsep=1.0pt,itemsep=1.0pt,leftmargin=5.5mm]
    \item \textbf{Place Bag}: Lift up a bag by the strap, and place it on a table. The bag's pose is varied.
    \item \textbf{Hang Hat}: Pick up a cap, and hang it on a rack. Caps' poses slightly vary across runs.
    \item \textbf{Tidy Tools}: Pick up a handled tool, place it in an open drawer, and close the drawer. We provide different tools, different initial tool poses, different drawers, and drawers to close.
\end{enumerate}

The combination of geometrically and visually diverse instance variation makes both learning manipulation skills and generalizing to new instances challenging.
\begin{figure}[h!]
  \centering
  \includegraphics[width=\textwidth]{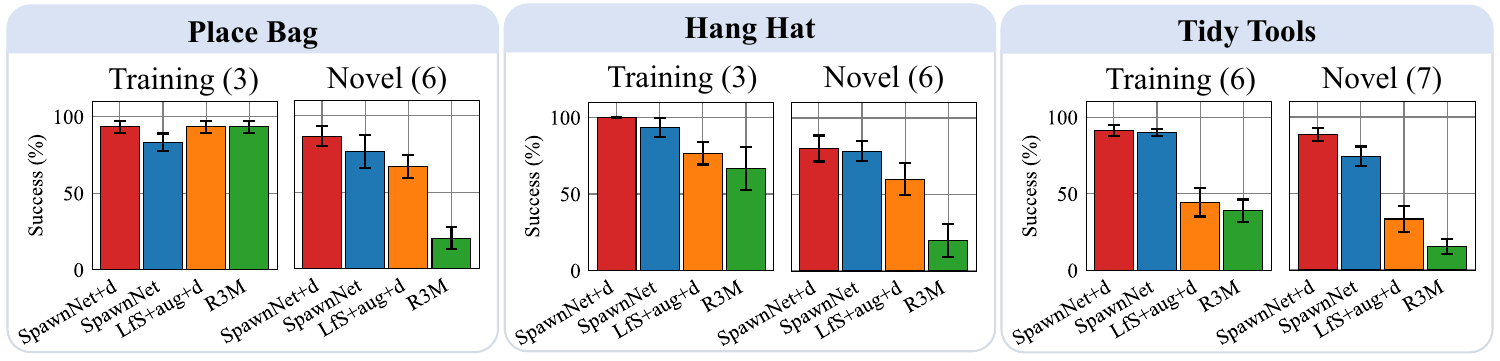}
  \caption{Real-world manipulation results on three tasks. All methods here use the same data augmentation. We evaluate each method on each instance over 5 trials (more than 30 trials on novel instances). We report the mean and standard error.} 
  \label{fig:real_results}
\end{figure}

\textbf{Results}. The results on training and held-out instances are shown in Figure~\ref{fig:real_results}. Numeric values and more details can be found in the appendix\footnote{Appendix link: https://xingyu-lin.github.io/spawnnet/files/appendix.pdf}. We observe that \algo~performs much better than baselines in both training and unseen instances; in unseen instances, the overall gap over the baselines becomes even larger than in simulation. We conjecture that this is because real-world images follow a closer distribution to the pre-training dataset compared to simulation. While LfS and R3M perform well on training occasionally, they are much worse at generalizing. For tasks with larger instance variations on training, such as Tidy Tools, both LfS and R3M fail to achieve 50\% success rates even on training while \algo~achieves greater than 80\% success. We hypothesize that this is because \algo~can better capture semantic variance across demonstrations on different instances through dense features. Additionally, by adding depth, \algo+d improves by as much as 15\% over \algo~on novel instances. This shows the flexibility of SpawnNet in incorporating different modalities.

\begin{figure}[ht]
  \centering
  \includegraphics[width=\linewidth]{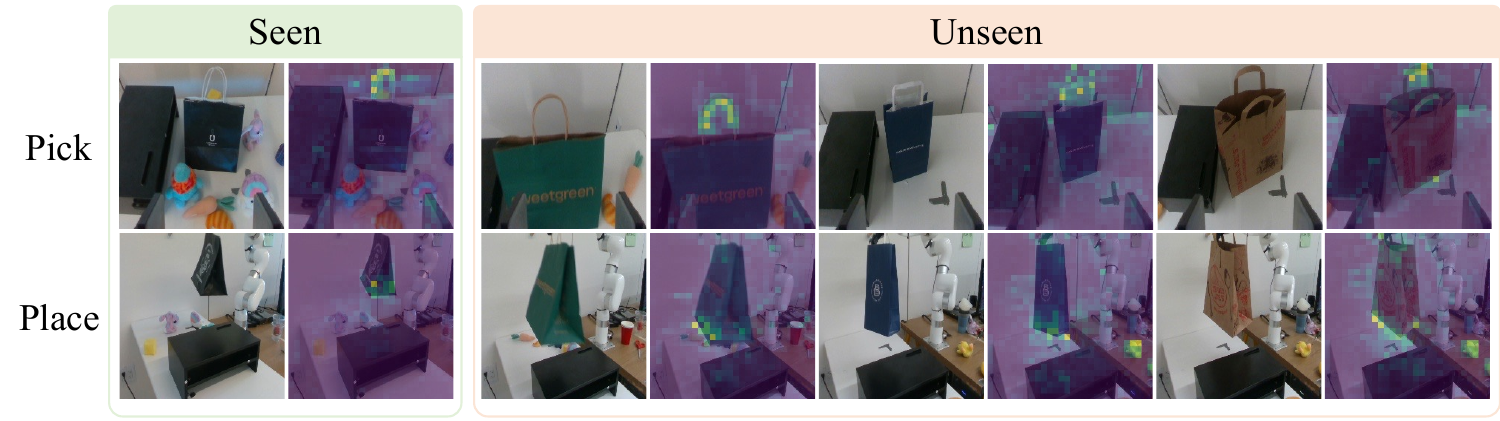}
  \caption{Visualization of the attention on the pre-trained features by the last adapter during the \textit{Place Bag} task. The attention focuses on important regions that are consistent across even unseen instances: the handles of the bags are highlighted when picking the bag and the bottom of the bags are highlighted when placing.}
  \label{fig:attn}
\end{figure}

\subsection{What does \algo~learn from the pre-trained network?} \label{sec:part_vis}
We provide more insights into how \algo's use of pre-trained features can help policy learning; to do so, we take a trained policy and visualize the norm of the features after the 1x1 convolutions and the nonlinear layers in the last layer of the adapter, as shown in Figure~\ref{fig:attn}. We see that important parts of the bags are highlighted, such as the handles of the bags when picking and the bottom of the bags when placing. These highlighted regions are consistent across time and across instances. They also generalize to unseen instances. As such, we believe that fusing the pre-trained network's layers enables better learning and generalization. Please refer to the website for more visualizations.

\subsection{Ablations} \label{sec:ablation}
We ablate SpawnNet's architecture using two tasks, Open Door (sim) and Place Bag (real):
\begin{itemize}[leftmargin=8mm]
    \item \textbf{Remove pre-trained features (-Pre-trained)}: To study how much we gain from the pre-trained representation, we zero-mask all pre-trained features.
    \item \textbf{Last pre-trained layer only (-Multiple)}: To study the importance of features from multiple layers, we only adapt the last pre-trained layer's features.
    \item \textbf{CLS token only (-Dense)}: To study the importance of dense features, we replace pre-trained layers' dense features [HxWxC] with the CLS token [1xC] tiled to the same dimensions.
\end{itemize}

\vspace{2mm}
\begin{wraptable}[8]{r}{.475\linewidth}
    \centering
    \vspace{-5mm}
    \begin{tabular}{l|ll}
        \toprule
        \textbf{Method}              & \textbf{Open Door} & \textbf{Place Bag} \\
        \midrule
        \textbf{Full Method}         & 59.6 & 86.7                  \\
        -Pre-trained                 & 52.9 \color{MyDarkRed}{(-6.7)} & -                      \\
        -Multiple                    & 58.3 \color{MyDarkRed}{(-1.3)} & 73.3  \color{MyDarkRed} (-13.3)                    \\
        -Dense                       & 52.5 \color{MyDarkRed}{(-7.1)} & 61.7 \color{MyDarkRed} (-24.9)                    \\
        \bottomrule
    \end{tabular}
    \caption{Ablations on Open Door and Place Bag; numbers in red indicate performance decrease compared to the full method. Removing dense features has the most impact.}
    \label{tab:ablation}
\end{wraptable}
Results are shown in Table~\ref{tab:ablation}. \textbf{Full Method} denotes SpawnNet+d. Notably, both removing dense features and pre-trained features entirely hurt performance, suggesting the importance of dense features for generalization. Adapting only the last pre-trained layer results in a slight decrease in performance, suggesting that spatial information from multiple layers is also helpful.

\vspace{3mm}
\subsection{SpawnNet with other pre-trained networks} \label{sec:other-networks}
\begin{wraptable}[12]{r}{.475\linewidth}
    \centering
    \vspace{-5mm}
    \begin{tabular}{l|ll}
        \toprule
        \textbf{Method}           & \textbf{Open Door} & \textbf{Place Bag} \\
        \midrule
        DINO                  &      44.6 & 11.7              \\
        R3M                       &   42.5  &  20.0               \\
        MVP                       &   43.8  & -                \\ \midrule
        SpawnDINO                 &   58.3 \color{MyDarkGreen}{(+13.7)}  & 86.7 \color{MyDarkGreen}{(+75.0)}                 \\
        SpawnR3M                   &  49.2 \color{MyDarkGreen}{(+6.7)}   & 81.7 \color{MyDarkGreen}{(+61.7)}                  \\ 
        SpawnMVP                     & 53.3 \color{MyDarkGreen}{(+9.5)} & - \\
        \bottomrule
    \end{tabular}
    \caption{Performance on the Open Door and Place Bag tasks with different pre-trained networks. The numbers in green show improvement over the pre-trained network. Using SpawnNet improves the performance of all pre-trained representations.}
    \label{tab:other-networks}
\end{wraptable}

We additionally experiment with initializing SpawnNet from other pre-trained backbones (\textbf{SpawnDINO}, \textbf{SpawnMVP}, and \textbf{SpawnR3M}), and compare them to pre-trained networks without SpawnNet (\textbf{DINO}, \textbf{MVP}, and \textbf{R3M}). These models vary broadly in terms of dataset, training objectives, and architecture. Again, we evaluate using the simulated Open Door and real Place Bag tasks. The results in Table~\ref{tab:other-networks} show SpawnNet is a general architecture that can improve the performance of any pre-trained network.

\begin{table}[h]
    \centering
    \vspace{5mm}
    \caption{Comparisons of performance with other transfer learning methods.}
    \label{tab:transfer_learning}
    \resizebox{\textwidth}{!}{
        \begin{tabular}{c|cccccc}
            \toprule
            Tasks & \multicolumn{1}{c}{SpawnNet} & \multicolumn{1}{c}{Frozen} & \multicolumn{1}{c}{Full Finetune} & \multicolumn{1}{c}{Last-layer Finetune} & \multicolumn{1}{c}{LORA Finetune~\cite{hu2021lora}} & \multicolumn{1}{c}{ProgressiveNet~\cite{rusu2016progressive}} \\ 
            \midrule
            Place Bag & $\mathbf{86.7 \pm 6.4}$ & $11.7 \pm 4.4$ & $8.3 \pm 4.4$ & $15.0 \pm 5.3$ & $28.3 \pm 6.4$ & $51.7 \pm 8.3$ \\ \hline
            Tidy Tool & $\mathbf{88.6 \pm 4.4}$ & $7.6 \pm 2.8$ & - & $26.7 \pm 4.5$ & $41.9 \pm 7.0$ & $43.8 \pm 6.1$ \\
            \bottomrule
        \end{tabular}
    }
\end{table}
\clearpage
\subsection{Comparison with Transfer Learning Methods}
\label{sec:transfer}
In this section, we additionally compare with other transfer learning methods. The results are shown in Table~\ref{tab:transfer_learning}. All methods are tested over the DINO pre-trained ViT for two real-world tasks. SpawnNet significantly outperforms other transfer learning methods across the board. Fine-tuning methods do not perform as well as SpawnNet, showing the difficulty of fine-tuning transformer models with limited demonstration data. Moreover, SpawnNet outperforms ProgressiveNet. The main difference is that SpawnNet uses a CNN for the control stream to take advantage of the inductive bias given limited demonstration data while ProgressiveNet uses symmetric architectures (i.e. ViT). 

\section{Limitations} 
Our experiments mainly study policy generalization under imitation learning settings, which requires the action distributions to be similar in training and evaluation tasks. For this reason, we only evaluate the policy on novel instances, without heavily extrapolating the pose of instances (see Appendix~C for details), as an out-of-distribution pose can cause a drastic change of motion. However, this can be addressed in an interactive learning setting. We leave incorporation with reinforcement learning algorithms to future work.

\section{Conclusions}
In this paper, we propose a simple and effective architecture to take advantage of pre-trained representations beyond simply freezing the representation. We show that \algo's use of the pre-trained representation not only improves performance on training instances but more importantly, offers better generalization to novel instances when compared to competitive learning-from-scratch and pre-trained baselines. Through systematic evaluation, we hope our paper can convince more people of the benefits of using pre-trained visual representations and boost further progress.

\acknowledgments{We would like to thank Youngwoon Lee, Xingyang Geng, Boyi Li, Qi Dou, and Jason Anderson for the fruitful discussions. We also thank Olivia Watkins, Yuqing Du and Oleh Rybkin for their feedback on the initial draft of the paper. This work was supported in part by NSF under the AI4OPT AI Center, the BAIR Industrial Consortium, and the InnoHK of the Government of the Hong Kong Special Administrative Region via the Hong Kong Centre for Logistics Robotics. }

\bibliography{bib/conference_macro,bib/references,bib/joso}
\clearpage

\newpage
\pagenumbering{arabic}
\renewcommand*{\thepage}{A\arabic{page}}
\appendix
\addcontentsline{toc}{section}{Appendix} 
\part{Appendix} 

\parttoc 

\section{Additional Experiments}

\begin{wrapfigure}[10]{R}{.33\linewidth}
  \vspace{-10mm}
  \centering
  \includegraphics[width=\linewidth]{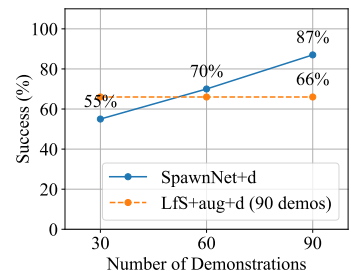}
  \caption{Comparing the number of training demos on \textit{Place Bag}. With as few as 60 demos, \algo~exceeds LfS+aug+d with 90 demos.}
  \label{tab:real_num_demos}
\end{wrapfigure}

\subsection{Real World: How much data is necessary?} 
Given \algo's success in the real-world experiments with little data, we aim to see how well our method generalizes given differing amounts of demonstrations. To do so, we additionally evaluate \algo+d with 30, 60, and all 90 demonstrations. For comparison, we additionally include LfS+aug+d with 90 demos; results are reported in Table~\ref{tab:real_num_demos}.


\algo+d performs surprisingly well with few demonstrations; it approaches the performance of the LfS+aug+d baseline with fewer than a third of the training demonstrations and exceeds it with two-thirds, demonstrating the effectiveness of dense features even with few demonstrations.

\label{sec:arch-design}
\section{Architecture Design Choices}
In this section, we describe specific choices in the architectures that we evaluate in our experiments.

\textbf{Encoders}. When comparing pretrained encoders, we control for similar parameter counts. MVP and DINO both use ViT-S (22M parameters), which has 12 layers. R3M uses ResNet-50 (23M parameters). For all encoders, we process each frame individually with the encoder, and concatenate representations across stacked frames and views before passing it into the MLP.

\begin{figure}[ht]
  \centering
  \includegraphics[width=.75\linewidth]{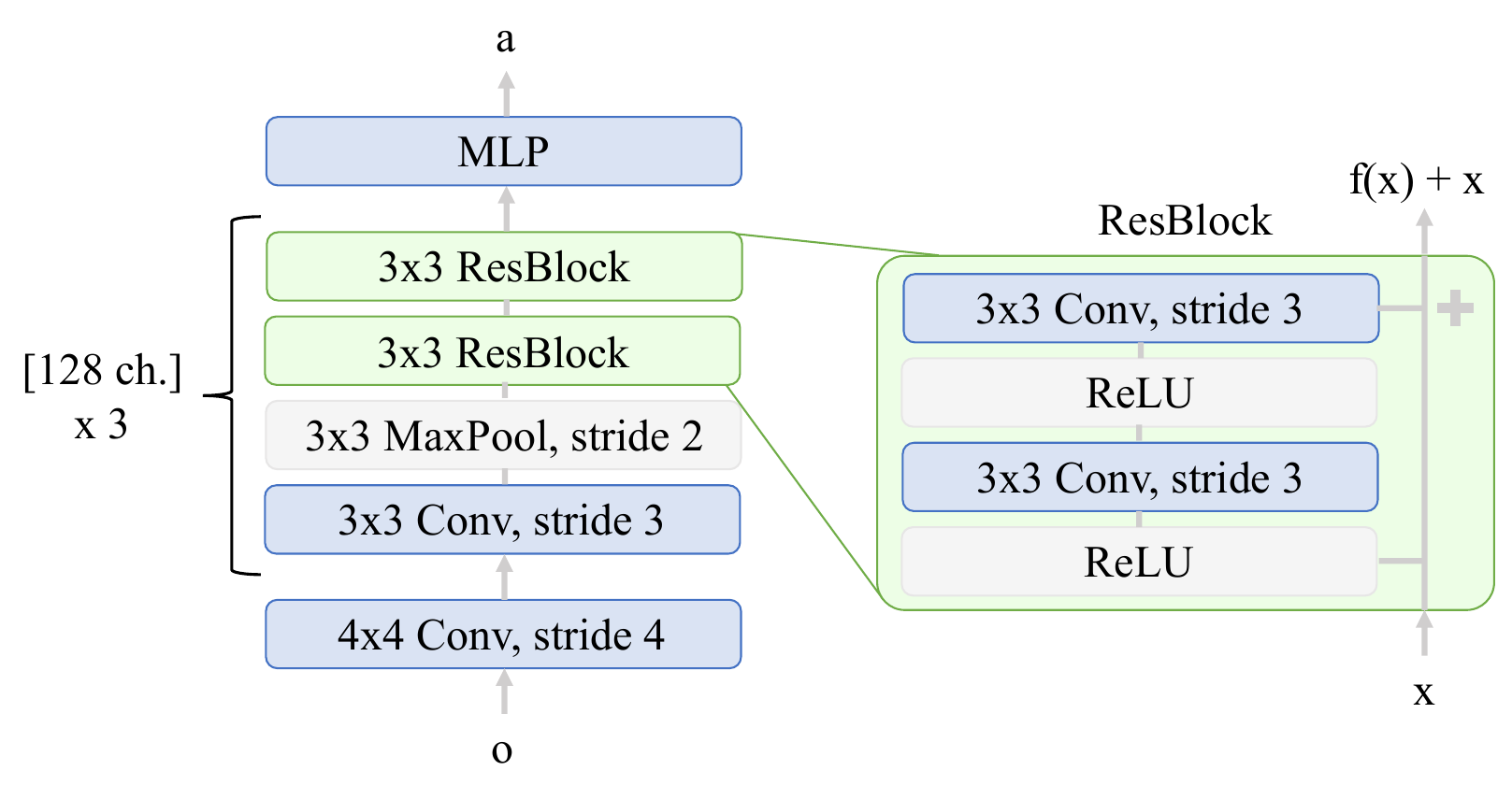}
  \caption{The convolutional encoder we consider for the LfS baseline. The initial 4x4 convolution transforms the input from its initial channel dimension to 128.}
  \label{fig:lfs_arch}
\end{figure}

We further expand on our learned encoder architectures:

\begin{itemize}[leftmargin=8mm]
    \item \textbf{Learning-from-scratch architecture}: Our LfS architecture follows the deep convolutional encoder described in \cite{espeholt2018impala}, with 128-channel 3x3 convolutions and 128-channel residual layers in each block. We detail this architecture in Figure~\ref{fig:lfs_arch}.
    
    \item \textbf{SpawnNet}: For SpawnNet, we use ViT-S/8 with a stride of 8, resulting in spatial attention features of shape [384, 28, 28]. SpawnNet uses three adapters, taking pre-trained features from the 6$^{th}$, 9$^{th}$, and 12$^{th}$ layers of the vision transformers respectively and mapping them to 64 channels (see Section~\ref{sec:arch}) Each adapted 64-channel feature is then concatenated to the current learned 64-channel feature before a 128-channel residual block.
\end{itemize}

\textbf{MLP}. We parameterize the MLP for all encoders using the same architecture, with hidden layers of size [256, 128]. The feature vector extracted from the encoder is flattened first before being passed in.

\textbf{Model Sizes and Inference Time}. We report the number of trainable parameters and inference time for models trained on xArm tasks in Table~\ref{tab:inference}. We note that \textit{Inference} is the real-time inference speed; \textit{Cached Inference} is the time taken for a forward pass with pre-calculated features (i.e. for training). Our LfS baseline has similar numbers of trainable parameters and cached inference speed as SpawnNet. Additionally, a SpawnNet backbone has approximately the same inference speed as a frozen pre-trained backbone.

\begin{table}[htbp]
    \centering
    \begin{tabular}{l|ccc}
    \toprule
    \textbf{Model}  & \textbf{Trainable Params (M)} & \textbf{Inference (ms)} & \textbf{Cached Inference (ms)} \\ \hline
    DINO            & 0.84                                 & 57.27                        & 0.27 \\
    Spawn-DINO      & 14.86                                & 59.17                        & 2.91 \\
    R3M             & 4.25                                 & 11.64                        & 0.16 \\
    Spawn-R3M       & 15.02                                & 14.15                        & 2.83 \\
    LfS             & 15.11                                & 2.47                         & 2.47 \\
    \bottomrule
    \end{tabular}
    \vspace{4mm}
    \caption{Inference times for different models. The increase in parameters between pre-trained and SpawnNet is from the use of dense spatial features instead of the CLS token.}
    \label{tab:inference}
\end{table}




\textbf{Data Augmentations}: Following \citet{hansen2022pretraining}, we consider random shift and random color jitter data augmentations. For simulation tasks, we only apply data augmentation to LfS+aug with $p_\textit{aug}=0.5$. For real tasks, we apply data augmentation to all methods with $p_\textit{aug}=0.5$. We provide psuedocode for our implementations below:
\begin{verbatim}
import torchvision.transforms as T

sim_aug = T.Compose([ # random shift
    T.Pad(5, padding_mode='edge'),
    T.RandomResizedCrop(size=224, scale=(0.7, 1.0))])

real_aug = T.Compose([ # random shift (no pad) and color jitter
    T.RandomResizedCrop(size=224, scale=(0.7, 1.0)),
    T.ColorJitter(brightness=0.3)])      
\end{verbatim}


\section{Details on Real World Tasks}
\label{sec: real_world}
We provide more details about the real-world tasks, including the total number of demos, the breakdown of demos per instance, and further details about the experimental setup.

\textbf{Place Bag}: 102 total demonstrations, with 34 demonstrations split across a red, black, and brown bag. Bags are placed in front of the robot, with variations in the x-y position (within a 1'x1' box) and the rotation (within a 90-degree range). The stand is kept fixed on the table. The table itself translates up and down within a 3" range, adding height variation as well.

\textbf{Hang Hat}: 95 total demonstrations, with 30 demonstrations on a teal hat, 33 demonstrations on a black hat, and 32 on a navy hat. Demonstrations grab above the bill of the hat, and hats are placed on a fixed stand with varying rotation within a 90-degree range. The table height remains fixed.

\textbf{Tidy Tools}: 90 total demonstrations, with 15 per drawer. We define a drawer as a level on the shelf, and leave some "levels" as novel instances; this tests the policy's ability to generalize learned features spatially. We additionally vary the tool being manipulated between two different handled tools and split these with 45 total demonstrations per tool across 6 different drawers. Tools are placed with a rotation within a 90-degree range inside of a 5"x5" box. Different drawers are placed with a rotation within a 45-degree range inside of a 5"x5" box. The table height remains fixed.

\textbf{Evaluation}. We place the novel instances within the training instances' pose variations as described above. Following \cite{karamcheti2023language}, we additionally award partial credit for tasks which consist of multiple manipulation skills; for example, in the \textit{Hang Hat} task, if the policy grasps the hat but is unable to hang it, we count the grasp as a success and the hang as a failure for a score of 0.5. The success rates are reported as the average success rate per instance. We perform 5 rollouts for each instance.


\section{Experimental Results}
We produce numeric tables for all experiments presented. 

\subsection{Numerical Results for Simulation Experiments}
The results in Table~\ref{tab:sim_num_results} correspond to the analysis presented in Section~\ref{sec:sim}.

\begin{table}[ht]
    \centering
    \begin{tabular}{l|cc|cc}
    \toprule
    \multirow{2}{*}{\textbf{Method}} & \multicolumn{2}{c|}{\textbf{Open Door}}                                & \multicolumn{2}{c}{\textbf{Open Drawer}}                              \\ \cline{2-5} 
                                     & \multicolumn{1}{c}{Train} & \multicolumn{1}{c|}{Val} & \multicolumn{1}{c}{Train} & \multicolumn{1}{c}{Val} \\ \hline
    SpawnNet                         & 87.3 $\pm$ 1.6                     & 58.3 $\pm$ 5.1                   & 85.7 $\pm$ 2.8                     & 61.1 $\pm$ 3.2                   \\
    LfS+aug                       & 90.5 $\pm$ 2.8                     & 53.3 $\pm$ 2.4                  & 81.0 $\pm$ 2.8                     & 61.7 $\pm$ 2.4                   \\
    R3M                              & 92.1 $\pm$ 4.2                     & 42.5 $\pm$ 5.1                   & 82.5 $\pm$ 3.2                     & 50.3 $\pm$ 1.8                   \\
    MVP                              & 68.3 $\pm$ 4.2                     & 43.8 $\pm$ 4.7                   & 58.7 $\pm$ 4.2                     & 30.0 $\pm$ 1.6                  \\ \hline
    PointCloudRL (expert)                     & 89.5 $\pm$ 0.0                     & 14.4 $\pm$ 0.0                   & 95.2 $\pm$ 0.0                     & 65.8 $\pm$ 0.0                  \\
    \bottomrule
    \end{tabular}
    \vspace{4mm}
    \caption{Numerical results for the simulation experiments.}
    \label{tab:sim_num_results}
\end{table}

\subsection{Real Experiments}
The results in Table~\ref{tab:real_num_results} correspond to the analysis presented in Section~\ref{sec:real_world}.
\begin{table}[ht]
    \centering
    \begin{tabular}{l|cc|cc|cc}
        \toprule
        \multirow{2}{*}{\textbf{Method}} & \multicolumn{2}{c|}{\textbf{Place Bag}} & \multicolumn{2}{c|}{\textbf{Hang Hat}} & \multicolumn{2}{c}{\textbf{Tidy Tools}} \\ \cline{2-7} 
                                         & Train              & Val                & Train              & Val               & Train              & Val                \\ \midrule
        SpawnNet+d                         & $93.3 \pm 3.8$     & $86.7 \pm 6.4$     & $100 \pm 0.0$     & $80.0 \pm 8.5$     & $91.1 \pm 3.6$     & $88.6 \pm 4.4$     \\
        SpawnNet                       & $83.3 \pm 5.6$     & $76.7 \pm 10.8$    & $93.3 \pm 6.1$     & $78.3 \pm 6.4$    & $90.0 \pm 2.4$     & $74.3 \pm 6.3$     \\
        LfS+aug+d                          & $93.3 \pm 4.2$     & $66.7 \pm 7.6$     & $76.7 \pm 15.2$     & $60.0 \pm 10.5$    & $44.4 \pm 9.4$     & $33.3 \pm 8.4$     \\
        R3M                              & $93.3 \pm 3.8$     & $20.0 \pm 7.1$     & $66.7 \pm 13.9$    & $20.0 \pm 10.8$   & $38.9 \pm 7.4$     & $15.2 \pm 4.8$     \\
        \bottomrule
    \end{tabular}
    \vspace{2mm}
    \caption{Numerical performance on the real world tasks. We report the average of the total success rate for each instance. The bar denotes standard error.}
    \label{tab:real_num_results}
\end{table}

\label{sec:visualizations}
\section{Pretrained Feature Visualization} Following the results presented in Figure~\ref{fig:attn}, we present more examples of the learned features from the adapter layers. More visualizations can also be found on our project website.

\begin{figure}[ht]
  \centering
  \includegraphics[width=\linewidth]{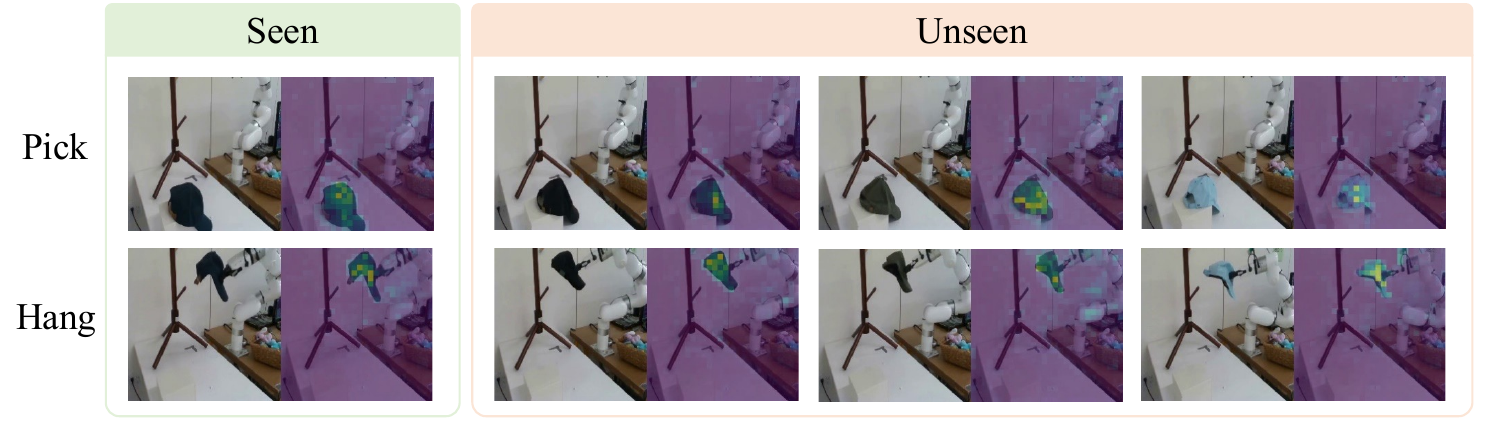}
  \caption{Visualized adapter features for the \textit{Hang Hat} task. When grasping the hat, the adapter highlights relevant parts of the hat, such as the brim and front. When hanging the hat, the adapter highlights relevant parts of the hat, such as the back edge.}
  \label{fig:hat_attn}
\end{figure}

\begin{figure}[ht]
  \centering
  \includegraphics[width=\linewidth]{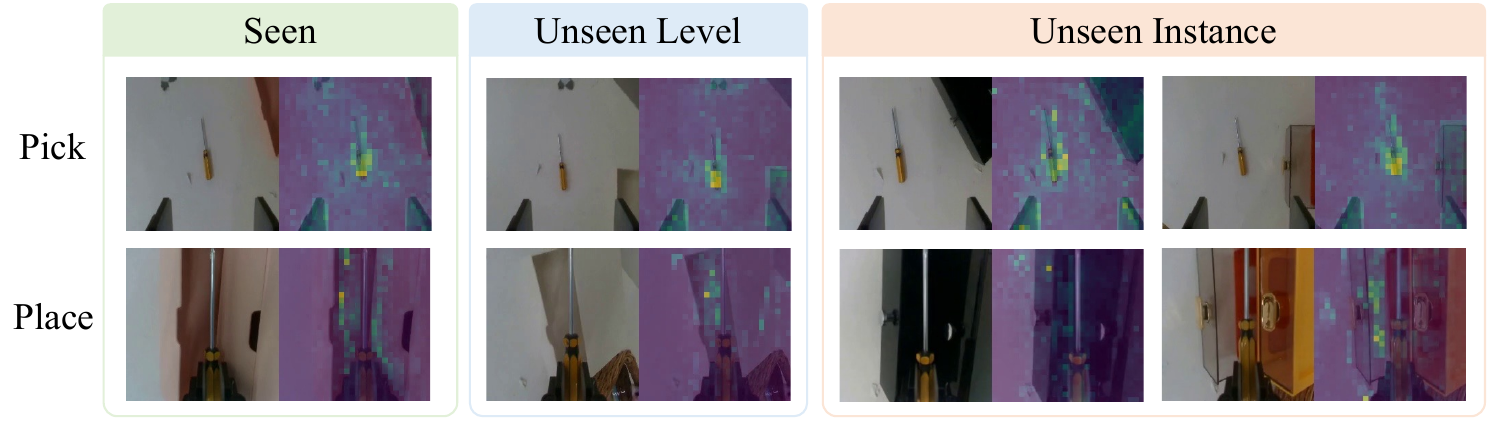}
  \caption{Visualized adapter features for the \textit{Tidy Tools} task. When grasping the tool, the adapter highlights its handle, even with novel drawers in the background. When placing the tool in the drawer, the adapter highlights the drawer's front edge.}
  \label{fig:drawer_attn}
\end{figure}

\end{document}